\documentclass[runningheads]{llncs}
\pdfoutput=1
\usepackage{graphicx}
\usepackage{orcidlink}
\usepackage{amsmath} 
\usepackage{amssymb}
\usepackage{eqparbox}
\usepackage{todonotes}
\usepackage{multirow}
\usepackage{MnSymbol}
\usepackage{mathtools}
\usepackage{enumitem}
\newlist{questions}{enumerate}{2}
\setlist[questions,1]{label=\textbf{EQ}\arabic*.,ref=EQ\arabic*}
\setlist[questions,2]{label=(\alph*),ref=\thequestionsi(\alph*)}
\begin{document}
\title{Intervening With Confidence: Conformal Prescriptive Monitoring of Business Processes\thanks{Supported by the European Research Council (PIX Project).}}
\titlerunning{Conformal Prescriptive Monitoring of Business Processes}
\author{Mahmoud Shoush\orcidlink{0000-0002-7423-9909} \and
Marlon Dumas\orcidlink{0000-0002-9247-7476}}
%
%
\authorrunning{M. Shoush and M. Dumas}
%
\institute{University of Tartu, Estonia \\
\email{\{mahmoud.shoush, marlon.dumas\}@ut.ee}}

%
\maketitle              


\begin{abstract} \label{sec:abstract}
Prescriptive process monitoring methods seek to improve the performance of a process by selectively triggering interventions at runtime (e.g., offering a discount to a customer) to increase the probability of a desired case outcome (e.g., a customer making a purchase). The backbone of a prescriptive process monitoring method is an intervention policy, which determines for which cases and when an intervention should be executed. Existing methods in this field rely on predictive models to define intervention policies; specifically, they consider policies that trigger an intervention when the estimated probability of a negative outcome exceeds a threshold. However, the probabilities computed by a predictive model may come with a high level of uncertainty (low confidence), leading to unnecessary interventions and, thus, wasted effort. This waste is particularly problematic when the resources available to execute interventions are limited. To tackle this shortcoming, this paper proposes an approach to extend existing prescriptive process monitoring methods with so-called \textit{conformal predictions}, i.e., predictions with confidence guarantees. An empirical evaluation using real-life public datasets shows that conformal predictions enhance the net gain of prescriptive process monitoring methods under limited resources.

\end{abstract}
\section{Introduction} \label{sec:introduction}

\textit{Prescriptive process monitoring (PrPM)} is a family of methods to optimize business processes by triggering runtime interventions with the goal of improving the percentage of cases that lead to a desired outcome~\cite{athey2017beyond}. For example, in a lead-to-order process, a PrPM method may recommend offering a discount (the intervention) to achieve a sales (desired outcome), while in an unemployment benefits assessment process, a PrPM system may allocate a problematic case to a senior case handler (intervention) to avoid an appeal (undesired outcome).

A range of PrPM approaches have been proposed in the literature~\cite{metzger2020triggering,bozorgi2021prescriptive,de2020design,fahrenkrog2021fire,shoush2021prescriptive}. These approaches consist of two components. The first component is a predictive model that estimates, for each ongoing case in a process, the probability that this case will end up in a desired ($dout$) or an undesired ($uout$) outcome ($out$).\footnote{Some approaches complement the prediction scores with a causal effect estimates~\cite{shoush2022intervene,bozorgi2021prescriptive,shoush2021prescriptive} to take into account the effectiveness of the interventions. However, the two components mentioned here are discernible also in these approaches.}
The second component is an \emph{intervention policy}, which determines if an intervention should be triggered for a given case, with the goal of optimizing a given gain function. This gain function generally considers the benefit of increasing the success rate (i.e.,\ more desired outcomes) as well as the cost of the interventions. 


A common weakness in existing approaches is that the intervention policy depends on predictions of case outcomes, which come with inherent uncertainty. Thus, a policy may trigger interventions based on misleading estimates. For example, the approaches in~\cite{teinemaa2018alarm,fahrenkrog2021fire} trigger interventions when the probability of the undesired outcome ($uout$) exceeds an empirically determined threshold, even when this probability is uncertain. This may lead to unnecessary interventions, which consume limited resources. In the abovementioned unemployment benefits process, the intervention involves escalating a case to a senior case handler. However, senior case handlers have a higher cost and limited capacity. Thus, escalating cases unnecessarily may lead to other cases not being escalated, even though the escalation would have been more effective for the latter.





In this paper, we enhance existing PrPM approaches with a third component, which determines if the uncertainty is low enough to assert that an intervention is warranted in the current state of a case. 
Specifically, we propose to use \textit{conformal prediction methods}~\cite{shafer2008tutorial} to obtain predictions with a confidence guarantee. 

The paper addresses the following research question: \textit{Does the use of conformal  predictions enhance the effectiveness, measured via a net gain function, of the interventions produced by an intervention policy in the context of a PrPM system?} 
To answer this question, we propose an architecture for injecting conformal predictions into a PrPM system in an environment where limited resources are available to execute the interventions. 
We conduct an empirical evaluation  using real-life public datasets to determine if and to what extent conformal predictions enhance the net gains produced by existing PrPM methods.

The next section discusses related work. Sect.~\ref{sec:approach} describes the proposed approach, while Sect.~\ref{sec:evaluation} presents and discusses the empirical evaluation. Finally, Sect.~\ref{sec:conclusion} concludes and gives future work directions.


\section{Related Work} \label{sec:bgrw}

A number of PrPM techniques have been proposed in the past decade. These techniques can be organized into three groups based on how the intervention policy prescribes interventions to improve a business value~\cite{kubrak2022prescriptive}. The first group focuses on control flow to recommend the next best action or activity to improve a pre-defined KPI~\cite{weinzierl2020prescriptive,DBLP:journals/corr/abs-2205-03219,de2020design}. The second focuses on a resource view to guide resource allocation decisions~\cite{sindhgatta2016context,park2019prediction}. Finally, the third focuses on triggering interventions to avoid or mitigate the effect of undesired outcomes and considers both control flow and resources perspectives~\cite{shoush2022intervene,shoush2021prescriptive,metzger2020triggering,fahrenkrog2021fire}.
The proposal in this paper belongs to the third group;  It seeks to learn policies for triggering runtime interventions to prevent undesired outcomes under limited resources. 

Most studies in this third group rely on predictive models trained on historical process execution data (event logs) to determine for which cases and when an intervention should be triggered. Fahrenkrog et al.~\cite{fahrenkrog2021fire} propose a PrPM approach that relies on predictions obtained from an outcome-oriented predictive model~\cite{teinemaa2019outcome} to determine if an intervention should be triggered for an ongoing case of the process. Specifically, this and similar methods trigger an intervention when the probability of an undesired outcome (herein denoted  $\mathbb{P}(uout)$) exceeds a threshold. This threshold is determined via a so-called empirical thresholding mechanism, which effectively tries out multiple thresholds over a subset of the event log to find the threshold that maximizes a reward function. This technique does not take into account the uncertainty inherent in prediction models.


Metzger et al.~\cite{metzger2020triggering} propose to use estimates of the reliability of predictions (calculated as per the method in~\cite{MetzgerF17}), in addition to a prediction score and other features, as input to an online reinforcement learning (RL) method. The RL method learns a policy that optimizes the percentage of cases that finish with a desirable outcome. 
However, the reliability estimates used by Metzger et al. do not come with confidence guarantees. Also, their approach learns a black-box policy since the reinforcement learner uses neural networks. Thus, the policy rules cannot be directly explained to a business user. Also, Metzger et al. use online RL, which means that the RL agent learns from trial-and-error in an actual operating environment until it converges into an optimal policy. In this paper, we seek to discover policies offline (i.e., based on past data).



In previous work~\cite{shoush2022intervene,shoush2021prescriptive}, we presented a PrPM technique that considers the tradeoff between triggering an intervention now versus later when resources are limited. This technique relies on estimates including the intervention effect (or conditional average treatment effect, i.e., $CATE$), \textit{total uncertainty} (determined as the entropy of the average prediction from an ensemble of machine learning (ML) classifiers), and the  $\mathbb{P}(uout)$. However, these uncertainty estimates do not come with confidence guarantees. This latter approach is used as a baseline in the empirical evaluation reported later in this paper.

\section{Approach} \label{sec:approach}
\begin{figure*}[!htb]
	\begin{center}
        \resizebox{0.8\textwidth}{!}{\includegraphics{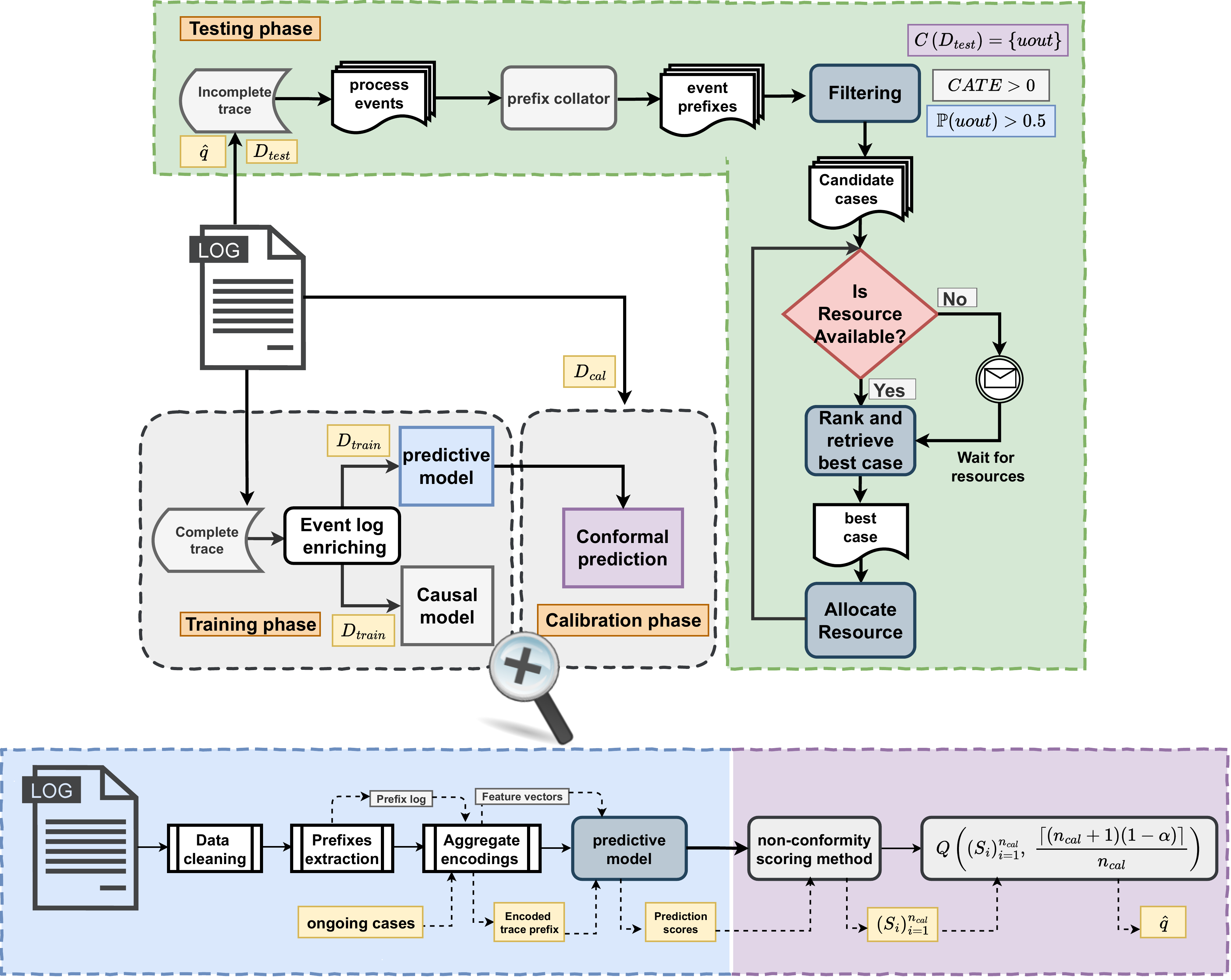}}
		\caption{An overview of the proposed approach. }
		\label{fig:approach}
	\end{center}
\end{figure*}

The proposed approach aims to construct an intervention policy. In line with existing ML approaches, the method consists of three phases, as illustrated in Fig.~\ref{fig:approach}: Training, Calibration, and Testing, which we discuss below in turn.





\subsection{Training phase}
In the training phase, we clean and enrich the event log and use it to train predictive and causal models. The predictive model is capable of estimating, for a given ongoing case, the probability that it will end in an undesired outcome ($\mathbb{P}(uout)$). The causal  model estimates the effect that an intervention would have if applied to an ongoing case, specifically, the increase in the probability of a positive outcome should the intervention be triggered (a.k.a.\ the Conditional Average Treatment Effect -- $CATE$). The details of how the predictive and causal models are trained are described in our previous work~\cite{shoush2022intervene,shoush2021prescriptive}. Below, we summarize this phase to make this paper self-contained.



\subsubsection{Event Log Enrichment}
This step includes \textit{data preparation, prefix extraction, enrichment, and prefix encoding}. In the data preparation step, we clean the event log from incomplete traces (given a definition of what makes a case “complete'') and we also remove events with clearly incorrect timestamps (outliers). We then extract prefixes according to length $K$ from every case to simulate real-life situations. Next, we enrich each prefix with additional attributes to help increase the accuracy of the predictive and causal models. Specifically, we add attributes related to the temporal context (date of the week, hour of the day of the most recent event in the case) as well as inter-case attributes (number of active cases and number of resources not currently actively working on a case). 
Finally, the extracted prefixes (or prefix log) are encoded into a fixed-size feature vector to train ML algorithms using the aggregate encoding method in~\cite{teinemaa2019outcome}.


The output of this step is a preprocessed dataset consisting of a set of tuples $(X_i, T_i, Y_i)$, where $X_i$ is a feature vector, including the original and  enriched features; $T_i$, is an intervention that could positively impact the outcome (herein denoted as $Y_i$). We then split this dataset into three folds $D_{train},  D_{cal}, D_{test}$ with $N = n_{train} + n_{cal} + n_{test}$ samples. We use each fold in the training, calibration, and testing phases, respectively.

\subsubsection{Predictive Model}
The purpose of the predictive model is to estimate, for any given prefix corresponding to an ongoing case, the probability that it will end in an undesired outcome. 
To train the predictive model, we apply a gradient-boosted tree algorithm, on the training fold $D_{train}$, with the goal of minimizing a loss function $\mathcal{L} (Y, \hat{Y})$, where $Y$ is the actual outcome, and $\hat{Y}(\{x\}_{i=1}^{n_{train}})$ is the predicted outcome.
The output from this step is a predictive model ($\hat{f}$) that produces a prediction score (probability) for the undesired outcome, $\mathbb{P}(uout)$, and thus also for the desired outcome, $\mathbb{P}(dout)$. 



\subsubsection{Causal Model}

The causal model estimates the causal effect of the intervention, i.e., the $CATE$. The $CATE$ is the number of percentage points by which the probability of the desired outcome $\mathbb{P}(dout)$ increases if we apply the intervention. For example, given a case of a lead-to-order process, which has a probability of 0.4 of ending in a desired outcome (a sales), a $CATE$ of 0.3 means that if the intervention is performed, the probability of a sales would increase to 0.7.


To estimate the $CATE$, we train a causal model that estimates the expected probability of undesired outcomes with, i.e., when $T=1$, and without, i.e., when $T=0$, applying the intervention. Then, the $CATE$ is the difference between these probabilities, given the current state of a case, represented by feature vector $X$.

\begin{equation}
    CATE = \mathbb{E}[(\mathbb{P}(out)_{T=0} - \mathbb{P}(out)_{T=1}) \mid X]   \hspace{2mm} \forall \hspace{2mm} out \in \{dout, uout\} 
    \label{eq:cate}
\end{equation}

\subsection{Calibration phase}

In this phase, we use an \textit{Inductive Conformal Prediction} (ICP) algorithm~\cite{shafer2008tutorial,tibshirani2019conformal} to produce predictions with a guaranteed level of confidence. 

ICP methods are model-agnostic, meaning they can be layered on top of any predictive modeling method. For example, we can use a random forest or a gradient boosting classifier to produce predictions, and then use an ICP method to transform the classifier’s outputs into a prediction with a confidence guarantee. Specifically, given a user-defined \textit{significance level} ($\alpha$) and a predictive model ($\hat{f}$), the ICP method returns a prediction set ($C$) that either contains a single outcome (e.g., $C=\{uout\}$) with confidence $1-\alpha$ (in which case the other outcome can be discarded at this confidence level), or $C=\{uout, dout\}$ if none of the outcomes can be discarded at the given significance level, or $C=\{ \}$ if
both outcomes can be discarded at the significance level. 


\begin{figure*}[!htb]
	\begin{center}
        \resizebox{0.8\textwidth}{!}{\includegraphics{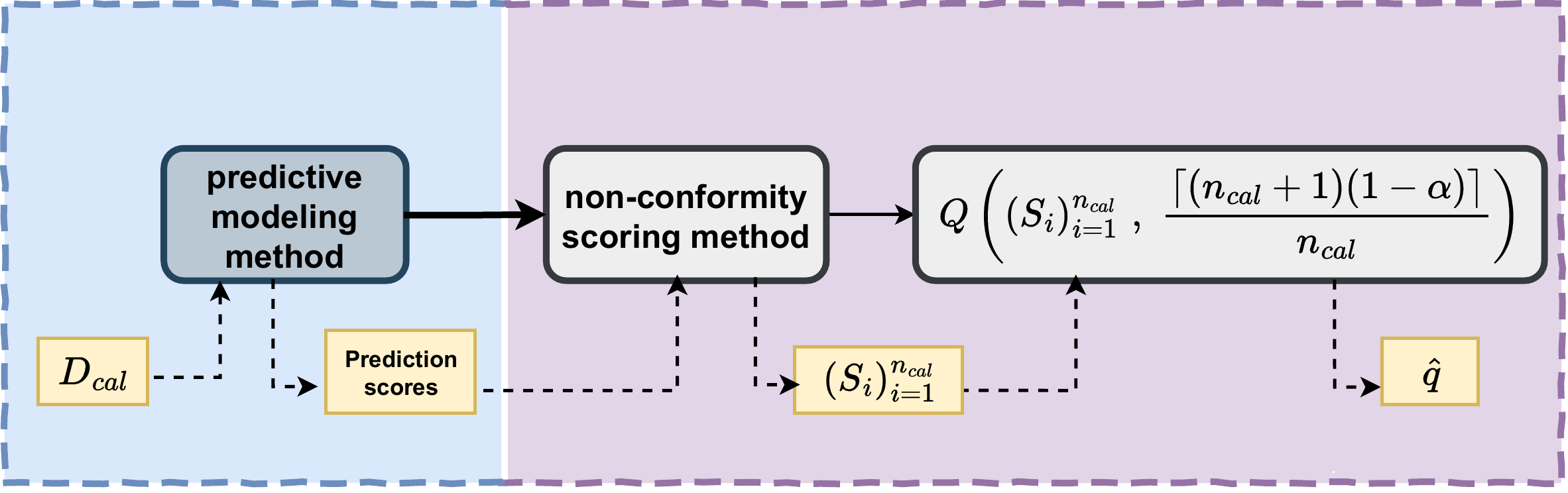}}
		\caption{The inductive conformal prediction method.}
		\label{fig:approach2}
	\end{center}
\end{figure*}

An ICP method works in two steps. In the first step, the ICP method calculates non-conformity scores ($S$) and a non-conformity quantile ($\hat{q}$). Fig.~\ref{fig:approach2} depicts the general approach to calculating $\hat{q}$. First, the predictive model is used to assign an outcome probability (i.e.,\ the prediction score) to each sample in the calibration data. Next, we give these predictive scores as input to a \emph{non-conformity scoring} method, which assigns a non-conformity score $s \in  (S_i)_{i=1}^{n_{cal}}$ to each sample. These non-conformity scores reflect how diverse a calibration sample is from training samples. For example, when the predictive model assigns lower prediction scores to the actual outcome in the calibration data, the non-conformity scores become high, meaning the uncertainty level is high. Then, $\hat{q}$ is calculated as a quantile of the resulting non-conformity scores, as per Eq.~\ref{eq:qhat}. The exact quantile used depends on the value of the significance level ($\alpha$).

\begin{equation}
    \hat{q} = Q\left((S_i)_{i=1}^{n_{cal}}, \hspace{2mm} \frac{\lceil (n_{cal}+1)(1-\alpha) \rceil}{n_{cal}}\right)
    \label{eq:qhat}
\end{equation}

In the second step, the value of $\hat{q}$ then determines which outcomes are included in the prediction set. 
Finally, and given $(X_{test}, Y_{test})$, i.e., an ongoing case at a given prefix, and the $\hat{q}$, ICP gives a \textit{prediction set ($C(X_{test})$)} with $1-\alpha$ confidence based on the marginal
coverage guarantee property~\cite{vovk1999machine}, as shown in Eq.~\ref{eq:confidence}. This property ensures that the actual outcome ($Y_{test}$) will be included in the prediction set ($C(X_{test})$) with $1- \alpha$ confidence. 
For example, assume the actual outcome is desired, i.e., $Y_{test} = dout$ and $\alpha=0.2$, then and according to Eq.~\ref{eq:confidence}.,  $ \{dout\} \in C(X_{test})$ with a confidence at least $80\%$.
The lower the $\alpha$, the higher the confidence. However, more confidence means the size of the prediction set increases to accommodate all possible outcomes. Still, in an outcome-oriented PrPM task, the most critical is getting $C(X_{test}) = \{uout\}$ to have more certainty that this case will end undesirably, then allocating resources efficiently. 

\begin{equation}
    \mathbb{P}(Y_{test} \in C(X_{test})) \geq 1-\alpha
    \label{eq:confidence}
\end{equation}

ICP methods differ in the way they calculate the non-conformity score, and in the way they use the non-conformity quantile $\hat{q}$ to determine the prediction set. Below, we describe the specific ICP methods we employ in our approach.

\subsubsection{Naive method}
Fundamentally, in the outcome-oriented task, the predictive model approximates $\mathbb{P}(Y =out \mid X = x )\hspace{1mm}  \forall \hspace{1mm}  out \in \{dout, uout\}$. For example, given an instance of a given case $x$, what is the probability of it belonging to $out$? Then we perform a naive calibration step by setting the non-conformity score ($S_n$) to be one minus the prediction score of the actual outcome, as shown in Eq.~\ref{eq:eq_naive}, to obtain $\{(s_i)\}_{i=1}^{n_{cal}}$. Then calculate $\hat{q}$ according to Eq.~\ref{eq:qhat}. 



\begin{equation}
    S_n = 1 - \hat{f}(X_{cal})_{{out}_{true}} \hspace{2mm} \forall \hspace{2mm} X_{cal} \in D_{cal}
    \label{eq:eq_naive}
\end{equation}
\begin{equation}
    C_n(X_{test}) = \{out: \hat{f}(X_{test})_{out} \geq 1 - \hat{q}\}
    \label{eq:eq_naive3}
\end{equation}
Then, the prediction set is constructed based on Eq.~\ref{eq:eq_naive3}, where $X_{test}$ is known, but $Y_{test}$  is not.  This means the prediction set will only include one outcome, desired or undesired, when the prediction score for one outcome satisfies the condition in Eq.~\ref{eq:eq_naive3}., and the other outcome does not. For example, when $\hat{q}=0.7$, the $\mathbb{P}(uout) =0.72$, and the $\mathbb{P}(dout)=0.28$, then the $C(X_{test})=\{uout\}$. Otherwise, the level of certainty about the prediction becomes insufficient to retain only one outcome and either include both or none.





\subsubsection{Outcome-balanced method}
This scoring ($S_{ob}$) method's principle is the same as the former; however, here, we perform the calibration step for each outcome separately to achieve outcome-balanced coverage, especially when the outcome of cases is imbalanced; thus, it guarantees~(\ref{eq:eq_ob1}) instead of (\ref{eq:confidence}). Hence, defining the non-conformity scores and non-conformity quantile for each outcome, as shown in Eq.~\ref{eq:eq_ob1}., means we stratify by the outcome.



\begin{equation}
     \mathbb{P}(Y_{test} \in C(X_{test}) \mid Y_{test}=out) \geq 1-\alpha, \hspace{1mm} \forall \hspace{1mm}  out \in \{dout, uout\}
    \label{eq:eq_ob1}
\end{equation}


\begin{equation}
    \hat{q}^{(out)} = Q\left((S_{i}^{(out)})_{i=1}^{n_{cal}(out)}, \hspace{2mm} \frac{\lceil (n_{cal}(out)+1)(1-\alpha) \rceil}{n_{cal}(out)}\right)
    \label{eq:eq_ob2}
\end{equation}

According to the outcome-balanced scoring method, the prediction set is determined by Eq.~\ref{eq:eq_ob3}., where we iterate over desired and undesired outcomes. Then it retains or not each outcome according to its quantiles. For example, assume  $\hat{q}^{(dout)}=0.7$, $\hat{q}^{(uout)}=0.4$, the $\mathbb{P}(uout) =0.3$, and the $\mathbb{P}(dout)=0.7$. Then the prediction set examines each outcome with its prediction score and $\hat{q}$. Hence, the $\mathbb{P}(uout) =0.3$ is not greater than  $1$ minus $0.4$; accordingly, the prediction set will discard the undesired outcome. Conversely, the $\mathbb{P}(dout) =0.7$ is greater than $1$ minus $0.7$; thus, the prediction set will retain the desired outcome, meaning $C(X_{test})=\{dout\}$ only.





\begin{equation}
    C_{ob}(X_{test}) = \{out: \hat{f}(X_{test})_{out} \geq 1 -\hat{q}^{(out)} \}
    \label{eq:eq_ob3}
\end{equation}

\subsubsection{Adaptive method}
Here, compared to previous methods ($S_n$ and $S_{ob}$) that consider only the prediction score for the actual outcome, this scoring method ($S_a$) considers all possible outcomes until the sum of their prediction scores exceeds the $1-\alpha$ confidence. Eq.~\ref{eq:eq_ad1}., shows how the non-conformity scores are calculated, where $\pi(x)$ is the permutation of all possible outcomes that orders $\hat{f}(X_{test})$ from the most likely outcome to the less likely. The next step is to compute $\hat{q}$ as (\ref{eq:confidence}),  and the prediction set is formed according to Eq.~\ref{eq:eq_ad2}.
\begin{equation}
    S_a =  \sum_{i=1}^{out} \pi(X_{cal})_i
    \label{eq:eq_ad1}
\end{equation}
\begin{equation}
    C_a(X_{test}) = \{out: S_a  \geq \hat{q} \}
    \label{eq:eq_ad2}
\end{equation}

Based on this scoring method, there is no empty prediction set because the prediction set will retain only one outcome when the level of certainty about it is high. Otherwise, it will retain both outcomes, but with different orders. Specifically, we add outcomes one by one to the prediction set until the sum of their prediction score exceeds the  $\hat{q}$. For example, assume $\hat{q}=0.8$, $\mathbb{P}(uout) =0.45$, and the $\mathbb{P}(dout)=0.55$. We first sort the prediction scores from the most likely to the least, e.g., $\mathbb{P}(dout)=0.55$, followed by  $\mathbb{P}(uout) =0.45$. Then we add the most likely outcome to the prediction set if its prediction score does not exceed $\hat{q}=0.8$, meaning $\mathbb{P}(dout)=0.55 < 0.8$. Next, we sum the next outcome in order to the previous one, and if their sum does not exceed the $\hat{q}=0.8$ will include it; otherwise, we stop and not adding any other outcomes to the prediction set. Since $ 0.45 + 0.55$ is greater than $\hat{q}=0.8$, the prediction set will not include the second outcome in the prediction set, thus $C(X_{test})=\{dout\}$.

In summary, for a given active case represented by a prefix of a trace, we can estimate $\mathbb{P}(uout)$ and $CATE$ using the models obtained from the training phase, and we can calculate the prediction set $C(X_{test})$ using the conformal model obtained from the calibration phase. The next step is to explain how to use these estimates to construct an intervention policy.

\subsection{Testing phase}
Here, we show how to operationalize the proposed approach during runtime. We assume that events for various ongoing cases keep coming constantly, and we first collate them into prefixes using a \textit{prefix collator}. Thus, we obtain a stream of trace prefixes, and we need to process each trace prefix to determine if an intervention should be triggered from the corresponding case. 

For each incoming trace prefix $(X_{test})$, we estimate $\mathbb{P}(uout)$, $CATE$, and $C(X_{test})$. These estimates are used first to \textit{filter} ongoing cases in order to identify ongoing cases that are candidates for an intervention and then to rank the candidate cases according to a gain function that takes into account the benefit of a case reaching a desired outcome, and the cost of the intervention. 

To identify candidate cases, we check three conditions. First, we check that $\mathbb{P}(uout)$ (probability of undesired outcome) is above a threshold. This threshold is determined empirically in order to maximize the gain. Second, we check that the prediction set consists only of the undesired outcome, i.e.,\  $C(X_{test}) = \{uout\}$. Third, we check that the intervention will have a positive effect, i.e.,\ $CATE > 0$.  Having identified the candidate cases for intervention, we pick the one that gives the highest gain. The gain is calculated as the gain we obtain each time we avoid a negative outcome ($C_{uout}$), multiplied by the $CATE$ (the increase in the probability of achieving this gain), minus the cost of the intervention $C_{in}$.

\begin{equation}
    gain = CATE * C_{uout} - C_{in}
    \label{eq:gain}
\end{equation}

$C_{uout}$ and $C_{in}$ are user-defined parameters, which are likely to vary from one process to another. As an example,  Tab.~\ref{tab:gainex} shows a cost and gain table for six case prefixes of an event log of an unemployment benefits process. Here, the undesired outcome is that the customer lodges an appeal of an unemployment benefits decision. 
When the cost of creating an appeal is greater than the cost of giving a discount, we accept to give the discount to achieve more gain, cf.\ case $C$ in the table. In contrast, when the cost of creating the appeal is less than the cost of giving the discount, we accept having the appeal, cf.\ case $E$. 

\begin{table}[hbpt]
\centering
\caption{An example of costs and gains.}
	\label{tab:gainex}
\resizebox{0.6\textwidth}{!}{

\begin{tabular}{ccccccc}
\hline
\hspace{0.2cm} $CaseID$ \hspace{0.2cm}& \hspace{0.2cm} $\mathbb{P}(uout) > \tau_{=0.5}$ \hspace{0.2cm} & \hspace{0.2cm} $CATE$ \hspace{0.2cm}&\hspace{0.2cm} $C(X_{test})$\hspace{0.2cm} &\hspace{0.2cm} $C_{uout}$ \hspace{0.2cm}& \hspace{0.2cm}$C_{in}$\hspace{0.2cm} & \hspace{0.2cm}$gain$\hspace{0.2cm} \\ \hline
A   & 0.52  & 5  & \{uout\} & 6     & 6   & 24   \\
B   & 0.54 & -1   & \{dout\} & -     & -   & -    \\
C   & 0.7  & 6   & \{uout\} & 10    & 5   & 55   \\
D   & 0.7 & 3   & \{\}     & -     & -   & -    \\
E   & 0.55  & 3   & \{uout\} & 2     & 12  & -6   \\
F   & 0.76  & 4   & \{uout\} & 10    & 5   & 35  \\\hline
\end{tabular}
}
\end{table}

Additionally, Tab.~\ref{tab:gainex} shows that we have six cases at the current moment, each with different $\mathbb{P}(uout)$, $CATE$ and $C(X_{test})$. According to the filtering step, we exclude $CaseID=B$ and $CaseID=D$, because the intervention has a negative effect for them. Moreover, there is no confidence that they will end in an undesired outcome, as their prediction set is empty. Now assume that there is currently only one available resource to make a phone call to give a discount to the complaining customer. In this scenario, we allocate the available resource to the case with the highest gain, cf.\ $CaseID=C$. 

To complement the gain function defined above, we also define a loss function to capture the loss of allocating a resource to a case that does not need an intervention (i.e. the intervention does not change the outcome). We define the loss as follows: $loss = C_{in} + CATE * C_{uout}$.




\section{Evaluation} \label{sec:evaluation}
To assess the usefulness and applicability of the proposed PrPM approach, we empirically investigate how conformal prediction improves the intervention policy at runtime to use the most profitable of the available resources in a way that increases the total gain. To this end, we examine the situation where resources are finite and compare the proposed approach in two stages. First, methods that either rely on predictions without confidence~\cite{fahrenkrog2021fire} or with an estimate of the uncertainty level~\cite{shoush2022intervene}. Second, methods that are predictive and causal~\cite{bozorgi2021prescriptive} as state-of-the-art baselines by discussing the following evaluation questions:




    \begin{questions}
        \item At which significance level ($\alpha$) does the number of cases in the prediction set which contains only undesired outcomes with confidence is maximized?\label{rq:rq1}
        
        \item To what extent does  conformal prediction improve the total gain?\label{rq:rq2}
    \end{questions}

\subsection{Datasets}
We experimented with two publicly available real-life event logs from the same domain, i.e., the banking industry, called \emph{ BPIC2017\footnote{\url{https://doi.org/10.4121/uuid:5f3067df-f10b-45da-b98b-86ae4c7a310b}} and BPIC2012\footnote{\url{https://data.4tu.nl/articles/dataset/BPI_Challenge_2012/12689204/1}}}. These logs represent the execution of a loan origination process. Tab.~\ref{tab:dataset} shows the key characteristics of these logs. We use them because both are big enough, w.r.t. the number of loan applications. Also, have a specific definition for \textit{desired}, e.g., when customers accept the offer and sign the contract, and \textit{undesired} outcomes, e.g., when the bank cancels the application or the customer refuses the offer. Furthermore, they have a clear notion of \textit{interventions} that may reduce the probability of undesired outcomes, e.g., making a second offer to customers.


\begin{table}[!htb]
\centering
\caption{The loan application process statistics.}
	\label{tab:dataset}
	\resizebox{0.8\textwidth}{!}{
\begin{tabular}{ccccccc}
\hline
dataset           \hspace{2mm}        & \hspace{2mm} \# applications   \hspace{2mm}      & \hspace{2mm}\begin{tabular}[c]{@{}c@{}}min\\ length\end{tabular} \hspace{2mm}& \hspace{2mm}\begin{tabular}[c]{@{}c@{}}max\\ length\end{tabular} \hspace{2mm}& \hspace{2mm}\begin{tabular}[c]{@{}c@{}}last\\ activity\end{tabular}      \hspace{2mm}       & \hspace{2mm}outcome  \hspace{2mm} & \hspace{2mm}\begin{tabular}[c]{@{}c@{}}intervention\\ activity\end{tabular}\hspace{2mm} \\ \hline
\multirow{2}{*}{BPIC2017} & \multirow{2}{*}{31,413} & \multirow{2}{*}{10}                                  & \multicolumn{1}{c|}{\multirow{2}{*}{180}}            & A\_pending                                                          & desired   & -                                                               \\ \cline{5-7} 
                          &                         &                                                      & \multicolumn{1}{c|}{}                                & \begin{tabular}[c]{@{}c@{}}A\_Canceled \\ A\_Declnied”\end{tabular} & undesired & Creat\_Offer                                                    \\ \hline
\multirow{2}{*}{BPIC2012} & \multirow{2}{*}{13,087} & \multirow{2}{*}{15}                                  & \multicolumn{1}{c|}{\multirow{2}{*}{175}}            & A\_Approved                                                         & desired   & -                                                               \\ \cline{5-7} 
                          &                         &                                                      & \multicolumn{1}{c|}{}                                & \begin{tabular}[c]{@{}c@{}}A\_Canceled \\ A\_Declnied\end{tabular}  & undesired & Creat\_Offer                                                    \\ \hline
\end{tabular}
}
\end{table}

The logs contain diverse case and event attributes. We use them in our experiments in addition to other extracted attributes, e.g., the number of sent offers, monthly loan interest, and temporal features, to enrich the logs. Then, we define outcomes according to each case's last activity and determine the intervention according to the \textit{Creat\_Offer} activity for cases labeled with undesired outcomes, as shown in Tab.~\ref{tab:dataset}. After that, we extract length prefixes at most the $90^{th}$ percentile to avoid too long cases. Finally, we apply an \textit{aggregate encoding} method that attains as much information from the original log and outperforms other encoding techniques according to previous work~\cite{teinemaa2019outcome}—the resulting fixed-size feature vector used as input to train the ML algorithms.  


\subsection{Experimental Setup}
The experimental setup to construct an intervention policy with the conformal prediction requires splitting the log temporally into three categories training ($60\%$), calibration ($20\%$), and testing ($20\%$). The training set is used to train predictive and causal models. In contrast, the calibration set is used to build the prediction set, and the testing set evaluates the intervention policy at runtime.  

We use a \textit{Gradient Boosting Decision Tree (GBDT)} algorithm, i.e., \textit{Catboost}~\cite{prokhorenkova2018catboost}, to train the predictive model that estimates the probability of cases likely to end undesirably, i.e., $\mathbb{P}(uout)$. Also, we use the \textit{Orthogonal Random Forest (ORF)} algorithm implemented in the \emph{EconMl}\footnote{\url{https://github.com/microsoft/EconML}} to train a causal model to estimate the $CATE$. We use these methods here, since previous work has shown that both GBDT~\cite{teinemaa2019outcome} and ORF~\cite{shoush2022intervene} methods achieve good results w.r.t accuracy in predicting undesired outcomes and estimating the intervention effect.

During runtime, ongoing cases are first filtered into candidates where the $\mathbb{P}(uout)>0.5$, $CATE>0$, and the $C(X_{test})=\{uout\}$. These estimates help decide which cases are more likely to end undesirably with confidence and can be persuaded via the intervention. Then we set the $C_{uout} =20$, relatively high, to $C_{in}=1$ to estimate the expected gain when allocating resources to candidates. 


We use different evaluation measures to evaluate the proposed approach. To evaluate the ICP methods, we use the \textit{area under the roc curve (AUC)} in addition to the \textit{F-score}. These metrics are reliable, especially when data is imbalanced, and not biased against one of the outcomes. Additionally, how many cases belong to the $C(X_{test})$ when it retains only the undesired outcome? Because our goal is to target cases that will  end undesirably with confidence. 


For the intervention policy under limited resources, we use the\textit{ total gain}, i.e., the sum of gains obtained from executing the intervention per available resource. Together with the accuracy of allocating resources \textit{(accuracy/resource)}, i.e., the number of cases we allocate resources to when needed, i.e., actually undesired cases, divided by the total number of cases we allocate resources to.

\subsection{Results}
We show the results of the proposed approach by first exploring the effect of the user-defined significance level ($\alpha$) on the prediction set~\ref{rq:rq1}. Then we analyze how the total gain improves when resources are finite according to the intervention policy that considers conformal prediction~\ref{rq:rq2}. 
\begin{figure*}[!htb]
	\begin{center}
        \centering\resizebox{0.7\textwidth}{!}{\includegraphics{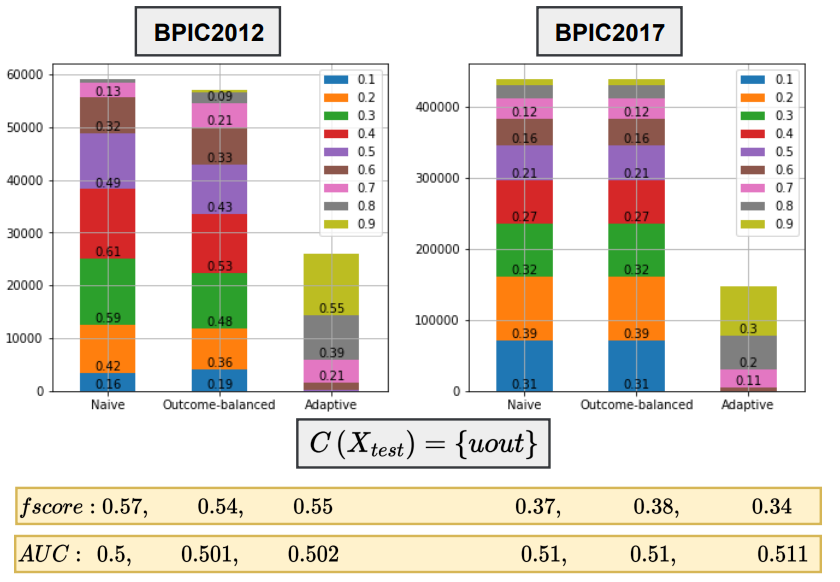}}
		\caption{The Histograms where $C(X_{test})=\{uout\}$ and the related metrics.}
		\label{fig:res1}
	\end{center}
\end{figure*}


In Fig.~\ref{fig:res1}. we show for each non-conformity scorning method how ($\alpha$) affects retaining undesired outcomes only in the prediction set to discuss ~\ref{rq:rq1}. We see clearly that for the naive and outcome-balanced methods, the number of cases that belong to the prediction set when retaining only undesired outcomes is maximized when $\alpha=0.4$ and $0.2$ for \textit{BPIC12} and \textit{BPIC17}, respectively. In contrast, in the adaptive method, $\alpha=0.9$ for both logs. This happens because of the way we construct the prediction set for each method. For the \textit{naive} and \textit{outcome-balanced} methods, the closer to $0$, the higher the $\hat{q}$. Accordingly, they are less conservative in adding a particular outcome to the prediction set, which is the opposite w.r.t the \textit{adaptive method}. Additionally, we observe that these significance levels achieve the highest \textit{F-score} and $AUC$ among all other significance levels, as shown in the  \emph{supplementary material}\footnote{\url{https://zenodo.org/record/7380386}}.  Hence, we use these levels to evaluate how conformal methods improve the \textit{total gain} and \textit{accuracy/resource}.

\begin{figure*}[!htb]
	\begin{center}
        \resizebox{0.7\textwidth}{!}{\includegraphics{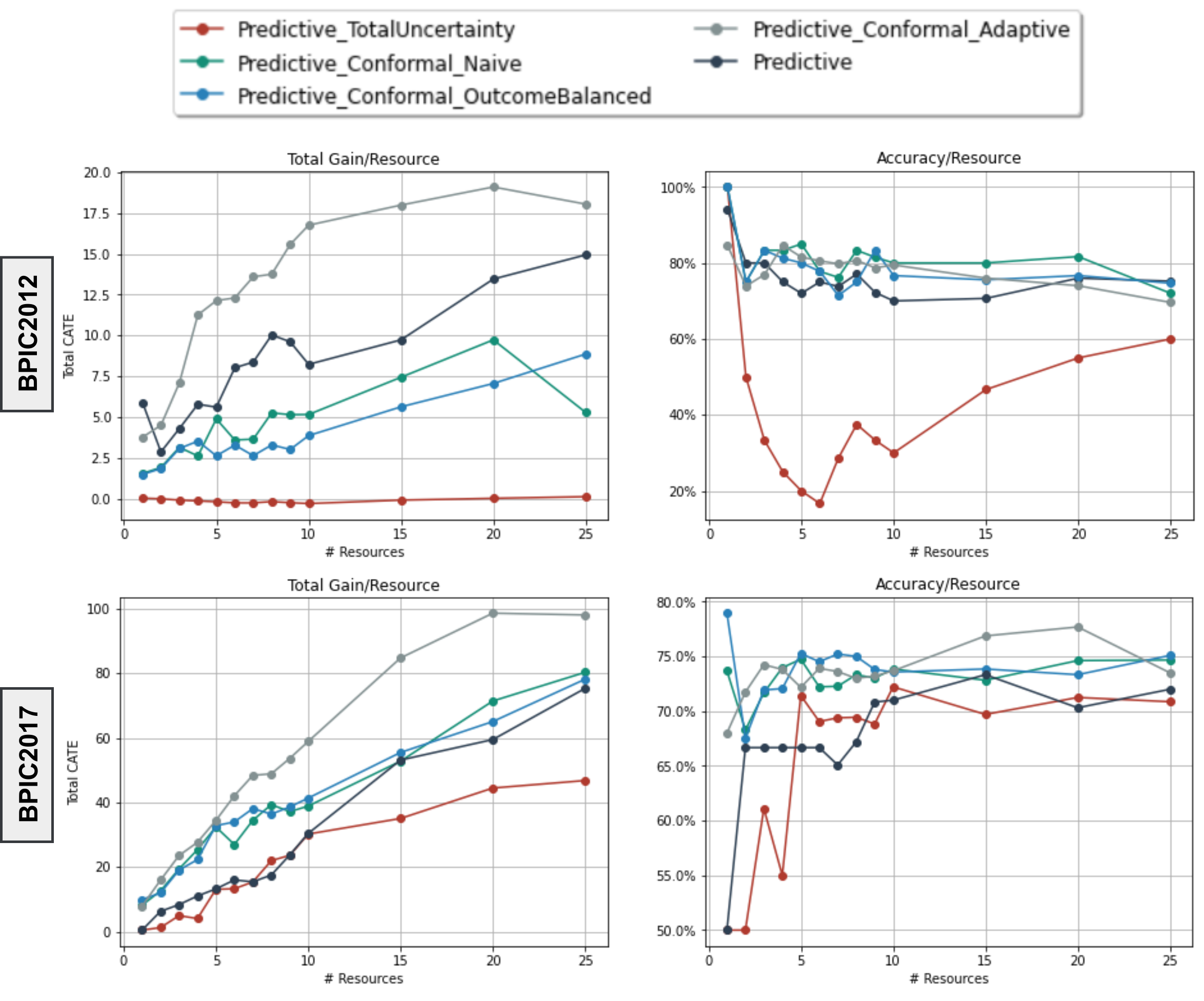}}
		\caption{Pure predictive VS predictive plus conformal. }
		\label{fig:res3}
	\end{center}
\end{figure*}

\begin{figure*}[!htb]
	\begin{center}
        \resizebox{0.7\textwidth}{!}{\includegraphics{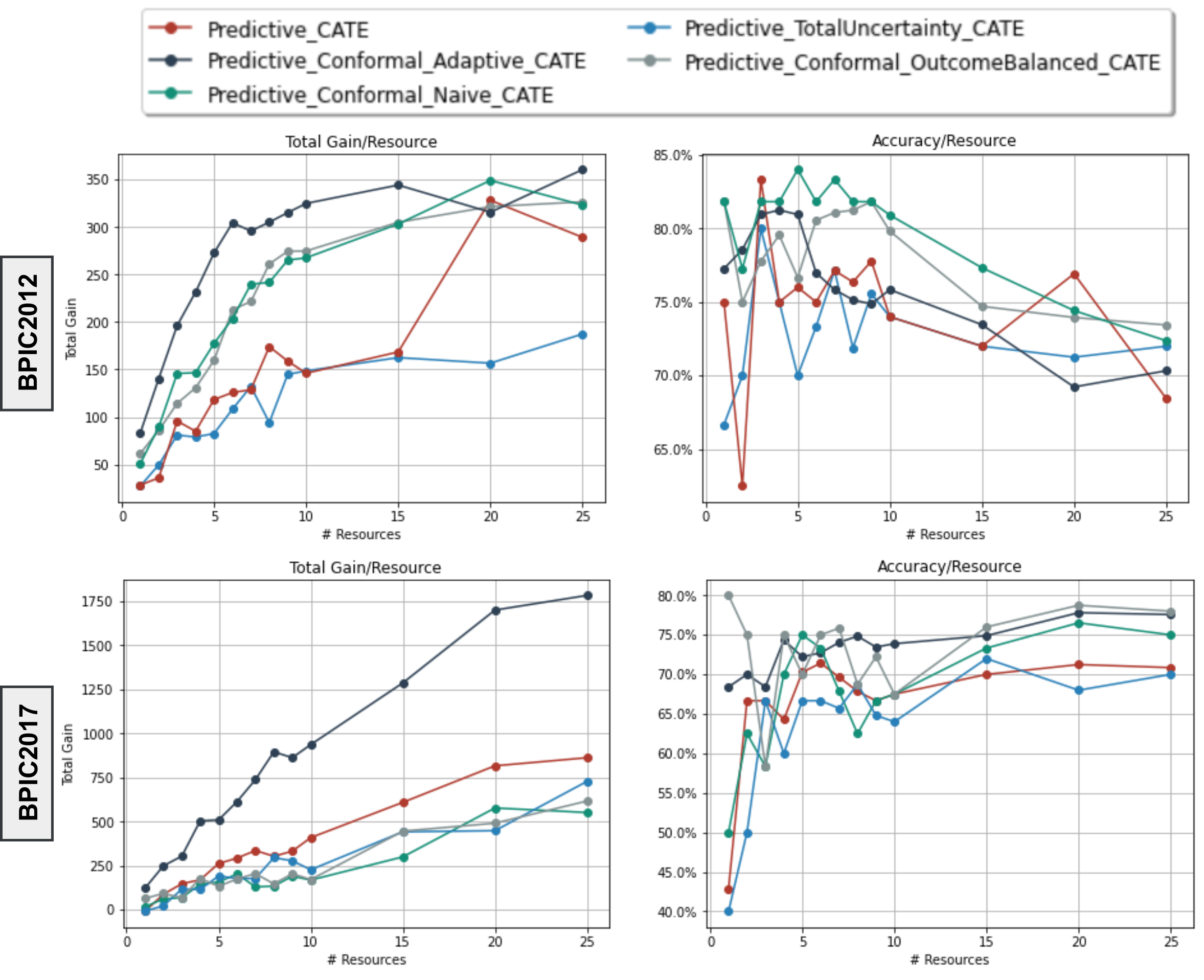}}
		\caption{Predictive plus CATE VS  Predictive plus CATE  plus Conformal}
		\label{fig:res4}
	\end{center}
\end{figure*}

To discuss~\ref{rq:rq2}, we first compare pure predictive methods that target cases when the $\mathbb{P}(uout)>0.5$~\cite{fahrenkrog2021fire}, with and without the $TotalUncertainty<0.75$~\cite{shoush2022intervene}, against predictive plus ICP (or conformal) when $C(X_{test})=\{uout\}$, see Fig.~\ref{fig:res3}. Here, we consider the gain equal to the benefits we earn from applying the intervention, i.e., CATE. Then, we compare predictive ($\mathbb{P}(uout)>0.5$ and the $TotalUncertainty<0.75$) plus $CATE$ (when it is above $0$)~\cite{bozorgi2021prescriptive} with predictive ($\mathbb{P}(uout)>0.5$) plus CATE plus conformal, see Fig.~\ref{fig:res4}.

For the \textit{BPIC2012} log, the \textit{total gain} (on the left-hand side) improves when we combine any conformal method with pure predictive in Fig.~\ref{fig:res3} and $CATE$ in Fig.~\ref{fig:res4}. In particular, when resources are minimal, with a remarkable \textit{accuracy/resource} compared to non-conformal methods. Also, the adaptive conformal method outperforms other methods w.r.t the total gain, and similar to other methods, w.r.t accuracy/resource. This is because the adaptive method's defined $\hat{q}$ is much higher than the naive and outcome-balanced methods; accordingly, more conservative in adding outcomes to the prediction set.

Moreover, when resources are not restricted, which is different from the situation in practice, we find that non-conformal methods achieve good gains with reasonable accuracy per resource as conformal methods. However, the conformal methods are more conservative since they constrain the allocation of resources.  

On the other hand, for the \textit{BPIC2017} log, the adaptive conformal method enhances the \textit{total gain} with high accuracy when resources are limited and relaxed. Conversely, when resources are limited, the naive and outcome-balanced methods achieve almost similar gains as non-conformal methods. However, still, all conformal methods outperform non-conformal methods w.r.t \textit{accuracy/resource}. 

The proposed PrPM approach outperforms baselines w.r.t the \textit{total gain} and \textit{accuracy/resource} according to Fig.~\ref{fig:res3} and Fig.~\ref{fig:res4}. Additionally, the  \emph{supplementary material}\footnote{\url{https://zenodo.org/record/7380386}} results show that our approach outperforms the work in~\cite{shoush2022intervene}. Then, building an intervention policy with limited resources using conformal prediction improves the performance of PrPM methods, thus, business processes.

\subsection{Threats to Validity}
We used two real-life event logs from the same domain in our evaluation. Thus, the evaluation lack generalizability. Therefore, the evaluation is preliminary and requires more experiments with other logs from other domains to be followed up. However, the field of business process mining needs more public logs with a clear notion of both outcome and intervention.

We assume that cases likely to end undesirably will be treated once, and the intervention effect is accurate and will directly minimize the probability of undesired outcomes. Still, there is a threat to ecological validity since cases can be treated more than once and with different actions or interventions. 

According to the literature, we used one type of algorithm for predictive, causal, and conformal methods. However, more experiments need to be done with other algorithms  to verify the validity of the developed intervention policy.  

\section{Conclusion and Future Work} \label{sec:conclusion}
We introduced a prescriptive process monitoring approach that designs  an intervention policy to learn for which cases we should trigger interventions and when to maximize the gain in a situation where resources are finite. The approach leverage a conformal prediction method combined with a predictive model to determine with confidence which cases are likely to end undesirably by creating a prediction set rather than a probability score for each outcome. Together with a causal model to estimate the impact of an intervention on the case outcome. 

At runtime, these estimates are given to a resource allocator that filters and ranks cases according to a set of rules to maximize the gain. Empirical evaluation shows that intervention policies with conformal instead of pure predictions outperform non-conformal methods, mainly when resources are finite.

Our proposal uses only one type of intervention and triggers it only once for cases. In practice, business cases can receive multiple interventions with different types. Thus, a direction for future work is to extend the current approach to tackle the problem of: which interventions can be triggered, for which cases, and when to maximize a business value, particularly when resources are limited. This gap can be modeled via a restless multi-armed bandit. Where cases are supposed to transition from one state to another based on the intervention.

\smallskip\noindent\textbf{Reproducibility.} The implementation and source code of the approach, detailed with instructions to replicate the evaluation, can be found at:  \url{https://github.com/mshoush/conformal-prescriptive-monitoring}.
\enlargethispage{0.5\baselineskip}

\bibliographystyle{splncs04}
\bibliography{mybibliography}
%




\end{document}